\documentclass[letterpaper, 10 pt, conference]{ieeeconf}  

\IEEEoverridecommandlockouts                              

\usepackage{xcolor}
\usepackage{soul}
\usepackage{graphicx}
\usepackage{hyperref}
\usepackage{caption}
\usepackage{subcaption}
\usepackage{array}
\usepackage{amssymb}
\usepackage{pgfplots}

\title{\LARGE \bf
LiLa-Net: Lightweight Latent LiDAR Autoencoder for 3D Point Cloud Reconstruction
}

\author{Mario Resino\textsuperscript{1}, Borja Pérez\textsuperscript{1}, Jaime Godoy\textsuperscript{1}, Abdulla Al-Kaff\textsuperscript{1} and Fernando García\textsuperscript{1}
\thanks{\textsuperscript{1}Authors are affiliated with Universidad Carlos III de Madrid, Department of Systems Engineering and Automation, Autonomous Mobility and Perception Lab (AMPL), Madrid, Spain
        {\tt\small \{mresino, boperezl, jgodoy\}@pa.uc3m.es},
        {\tt\small \{akaff, fegarcia\}@ing.uc3m.es}}
}

\usepackage{amsmath}
\usepackage{threeparttable}

\begin{document}

\maketitle
\thispagestyle{empty}
\pagestyle{empty}

\begin{abstract}

This work proposed a 3D autoencoder architecture, named LiLa-Net, which encodes efficient features from real traffic environments, employing only the LiDAR's point clouds.  For this purpose, we have real semi-autonomous vehicle, equipped with Velodyne LiDAR. The system leverage skip connections concept to improve the performance without using extensive resources as the state-of-the-art architectures. Key changes include reducing the number of encoder layers and simplifying the skip connections, while still producing an efficient and representative latent space which allows to accurately reconstruct the original point cloud. Furthermore, an effective balance has been achieved between the information carried by the skip connections and the latent encoding, leading to improved reconstruction quality without compromising performance. Finally, the model demonstrates strong generalization capabilities, successfully reconstructing objects unrelated to the original traffic environment.

\end{abstract}

\section{Introduction}

\begin{figure*}[tbph]
    \centering
    \includegraphics[clip, trim={0cm 19cm 0cm 4cm}, width=\linewidth]{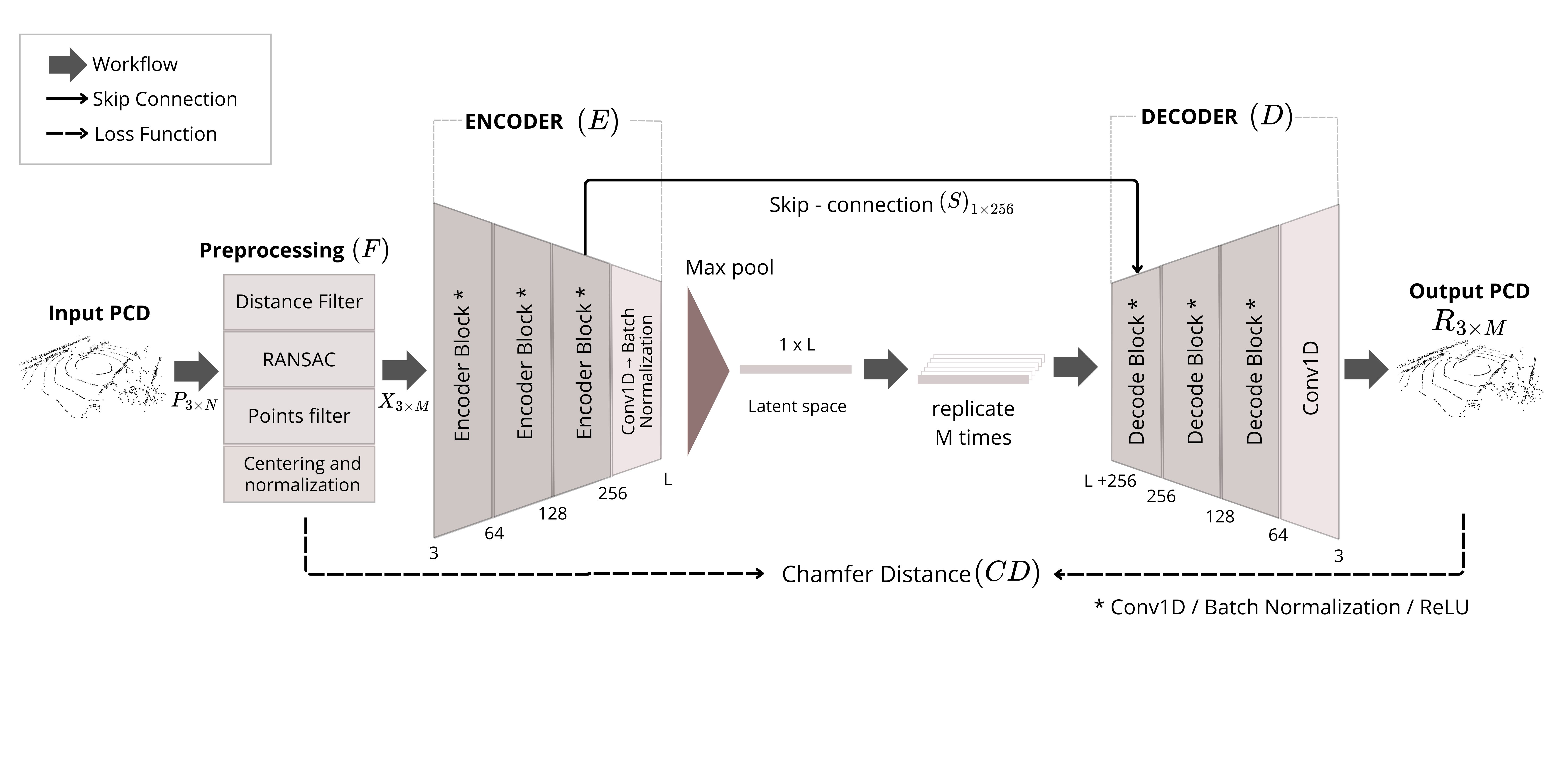}
    \caption{Proposed LiLa-Net Architecture: point clouds are first preprocessed ($F$). Then, multiple encoder layers ($E$) progressively reduce the information to the latent space dimension ($L$). For reconstruction, the latent representation is concatenated with the skip connection features ($S$) at the lowest level, followed by a series of decoder layers ($D$) to obtain the final reconstruction. The loss function is the Chamfer Distance between the reconstructed output and the preprocessed point cloud.}
    \label{fig:pipeline}
    \vspace{-0.9em}
\end{figure*}
The perception and understanding of the environment in autonomous vehicles remain a significant challenge, involving both hardware components - such as 2D/3D sensors, control units, proximity detectors, and audio sensors - and software modules, including path planning, obstacle detection, environment classification, and automatic control. Among these technologies, LiDAR sensors have emerged as a key tool for capturing accurate and detailed 3D representations of traffic environments. However, the high volume of data generated by point clouds creates the need for models that can efficiently extract meaningful features while minimizing computational and memory demands.

Recent advances have explored Transformer-based architectures for 3D point cloud processing~\cite{lu2022transformers}, leveraging attention mechanisms to capture spatial dependencies. While effective, these methods are often computationally expensive and resource-intensive, limiting their practical deployment in real-time systems.

To address these limitations, this work introduces LiLa-Net, a lightweight end-to-end framework for point cloud feature extraction and reconstruction. LiLa-Net learns compact and expressive latent representations that preserve the structural features of scenes, enabling efficient compression and consistent reconstruction with minimal error.

Furthermore, after validation in traffic-focused datasets, LiLa-Net demonstrated strong adaptability when applied to unrelated objects, showing promising generalization and fine-tuning capabilities even with limited additional data.

The key contributions of this work are as follows: $i)$ the proposal of LiLa-Net, a novel point cloud autoencoder framework that directly operates on sparse 3D points, avoiding voxelization or intermediate representations; $ii)$ the ability to efficiently handle large and dense point clouds, enabling processing at the scene level rather than on isolated objects; $iii)$ demonstration of effective compression and reconstruction of complex traffic environments collected from moving vehicles; $iv)$ evidence of strong generalization of the learned latent representations, achieving robust classification performance on entirely different datasets without retraining; and $v)$ elimination of pretraining or masking strategies, resulting in a simpler, faster, and more direct training pipeline.

The remainder of this paper is organized as follows. Section~\ref{sec:related_works} reviews the state-of-the-art (SOTA) methods in the field. Section~\ref{sec:methodology} describes the proposed framework, detailing its encoder–decoder architecture and the integration of skip connections. Section~\ref{sec:experiments} presents the experimental results, including an analysis of different architectural configurations, as well as qualitative and quantitative evaluations across various tasks, such as reconstruction and cross-dataset classification, in comparison with other SOTA approaches. Finally, Section~\ref{sec:conclusions} summarizes the key findings of this work.

\section{Related Works}\label{sec:related_works}

Recent advances in 3D point cloud processing have focused on the development of architectures capable of learning meaningful latent representations directly from raw data, effectively overcoming the limitations of earlier voxel or image-based approaches~\cite{liu2025voxelbasedpointcloudgeometry,tang2024sparseocc}. The dominant paradigm driving this progress is self-supervised learning (SSL)~\cite{sauder2019selfsuperviseddeeplearningpoint,achituve2022selfsupervisedlearningdomainadaptation}, which enables large-scale training without the need for manual annotation.

Early SSL approaches relied on pretext tasks, such as reconstructing point clouds that had been shuffled or partially deformed, thereby forcing the network to capture the underlying geometric structure of the object. While effective, these methods have recently been outperformed by masked modeling strategies, inspired by Masked Autoencoders (MAE)~\cite{jiang2023maskedautoencoders3dpoint, pang2022maskedautoencoderspointcloud} from other domains.

The masked modeling paradigm divides the point cloud into patches, masks a high percentage of them, and trains an asymmetric Transformer-based autoencoder to reconstruct the missing parts. This approach offers significant computational efficiency, as a powerful encoder processes only the visible patches, while a lighter decoder reconstructs the occluded regions. Models such as Point-MAE~\cite{pang2022maskedautoencoderspointcloud}, Point-BERT~\cite{yu2022pointbertpretraining3dpoint}, and hierarchical variants like Point-M2AE~\cite{zhang2022pointm2aemultiscalemaskedautoencoders} have demonstrated remarkable ability to learn robust and transferable features, which can be fine-tuned for downstream tasks, including classification and segmentation.

In parallel, generative modeling for point clouds has seen substantial progress. Classical autoencoders, such as FoldingNet~\cite{yang2018foldingnetpointcloudautoencoder}, introduced innovative decoders that “fold” a 2D grid to reconstruct 3D objects, while Variational Autoencoders (VAEs)~\cite{achlioptas2018learningrepresentationsgenerativemodels, han2019multianglepointcloudvaeunsupervised} enabled the creation of probabilistic latent spaces well-suited for shape interpolation. Although, Generative Adversarial Networks (GANs)~\cite{wu2017learningprobabilisticlatentspace, khan2019unsupervisedprimitivediscoveryimproved} played an important role in early research, diffusion models have now emerged as the state-of-the-art for high-fidelity generation. These models generate highly realistic 3D shapes by reversing a noise-injection process. Hybrid architectures, such as DiffPMAE~\cite{li2024diffpmaediffusionmaskedautoencoders}, combine the efficiency of MAE-based encoders with the generative power of diffusion, establishing a new benchmark in quality.

In summary, recent research highlights two main directions: methods designed to learn compact and structured latent spaces for effective representation learning, and generative approaches, such as diffusion models, that have achieved remarkable progress in 3D object synthesis. Both lines emphasize the importance of capturing geometric and semantic essence of 3D data, which continues to drive advancements across a wide range of downstream applications.

\section{Methodology}\label{sec:methodology}

The following sections describe the proposed methodology LiLa-Net, illustrated in Fig.~\ref{fig:pipeline}. The process begins with data preprocessing ($F$), after which the processed point cloud ($X$) is passed through the encoding ($E$) to extract the most relevant features and generate a compact latent representation. This latent vector, together with a skip connection ($S$) from the encoder ($E$), is then fed into the decoding block ($D$) to reconstruct the final point cloud ($R$) and evaluate it with the Chamfer Distance ($CD$) metric. Each stage of the pipeline is described in detail in the following subsections.

\subsection{Data Adquisition}
To conduct this study, a proprietary dataset was collected at the AMPL laboratory of Universidad Carlos III de Madrid using our recording platform, an updated version of Atlas platform~\cite{ATLASPaper}. From this dataset, only the point clouds with spatial information ($P_{3\times N}$) corresponding to complete sequences were extracted, which were captured using a Velodyne VLP-32C LiDAR sensor around the Center for Innovation in Entrepreneurship and Artificial Intelligence (C3N-IA) at the UC3M technological Park.

\subsection{Pipeline}

\subsubsection{Preprocessing}
The point clouds from this dataset must be preprocessed before being fed into $E$ due to the large proportion of points that do not contribute relevant scene information. For instance, ground points-which can dominate the input and undesirably influence the encoder’s attention. To address this, the classical RANSAC~\cite{fischler1981random} algorithm is employed to detect and remove ground points.

In addition to ground removal, a horizontal range filter is applied by defining a cylindrical region around the sensor. Points falling outside this parametric radius (ranging from 15 to 200 meters) are discarded. The effect of varying this radius is analyzed in the experimental results in Section ~\ref{subsec:cloud_size}.

Finally, to match the encoder's input requirements, the point cloud is randomly downsampled until get $M$ number of points ($X_{3\times M}$).

\subsubsection{Encoder}

From $X_{3\times M}$, $E$ is responsible for extracting a compact and highly rich feature representation, capturing its essential geometric features. It takes as input a tensor of shape $B\times3\times M$, with batch size $B$. $E$ is composed of a sequence of shared 1D convolutional layers, which operate independently on each point. These layers progressively increase the feature dimensionality from 3 to the desired latent space size ($L$). Each convolutional layer is followed by a Batch Normalization and ReLU activation to improve convergence and introduce non-linearity.

After the final convolution, a global feature vector is obtained by applying a max-pooling operation over the point dimension. This operation aggregates the most prominent feature across all points, resulting in a fixed-length latent vector of shape $1\times L$. 

\subsubsection{Latent Feature Space}

From the latent vector with $L$ size, the framework encodes the most relevant information required for reconstructing the original point cloud from the bottleneck. This process yields a latent space that captures the global 3D structure of the scene and the semantic context of its components, while discarding less informative content. The resulting representation has a fixed size of $1\times1024$ dimensions. In this way, the architecture can be trained to represent 3D point maps through a fixed-dimensional latent vector, invariant to density variations, point ordering, or minor scene deformations.

These latent representations have proven highly useful not only for reconstruction but also for a wide range of downstream tasks, including classification, clustering, retrieval, and generative modeling~\cite{Hartman2024SelfSupervised, Meng2022TopicDiscovery, Lan2024GaussianAnything}.

\subsubsection{Skip Connection}

In addition to the latent space, part of the information required for reconstructing the point cloud is carried through $S$, which transfers features extracted from a single encoder layer directly to the corresponding decoder stage
Based on the experiments described in Section ~\ref{subsec:skip_relevance}, the framework was designed to retain only the skip connection from the last encoder layer. This choice ensures that the reconstruction relies primarily on the latent space while preserving the minimal complementary information needed for a good reconstruction. As a result, the latent representation becomes richer and more informative.

\subsubsection{Decoder}

Once all the information from the original point cloud has been encoded and the features have been successfully extracted, the final module of our framework is $D$, responsible for transforming a global feature vector and the skip connection features back into a set of 3D coordinates representing the reconstructed point cloud $R_{3\times M}$.

In our architecture, $D$ is implemented as a sequence of shared 1D convolutional layers with kernel size 1, each followed by Batch Normalization and a ReLU activation, as in $E$. These layers progressively refine the feature maps until reaching the final output dimensionality, which corresponds to the desired number of 3D points.
\section{Experiments and Discussion}\label{sec:experiments}
Building on LiLa-Net, a series of experiments were conducted to evaluate the framework’s performance and to refine its components. The goal was to optimize each module, ensuring a robust reconstruction of the automotive environment while producing a high-quality latent space that effectively captures the scene’s structure and semantic content.

\subsection{Dataset}
First, a study was carried out to collect multiple recordings with varying characteristics using our recording platform at the Autonomous Mobility and Perception Lab (AMPL) of Universidad Carlos III de Madrid, Spain. A total of 4,955 point clouds were initially collected. Nevertheless, to assess the robustness of the proposed architecture, we performed experiments using different maximum range thresholds and varying the number of input points $N$. This analysis allowed us to test the feasibility of reconstructing larger point clouds in terms of both spatial extent and point density. After that $P$ is passing through $F$, which provides $X$ as an output and splitted into 4,460 ($\sim90\%$) point clouds for training and 495 ($\sim10\%$) point clouds for testing.

To further evaluate the generalization capability of our approach and to provide a comparison with existing methods, experiments were also conducted on the publicly available ModelNet10 and ModelNet40 datasets~\cite{wu20153d}, which consist of numerous point clouds of specific objects and are primarily used for classification tasks. Following the standard protocol, the official training and testing splits provided with each dataset were employed.

This combination of a proprietary dataset—captured under realistic automotive conditions—and publicly available benchmarks allows us to assess both the domain-specific reconstruction capabilities of our autoencoder and its transferability to more general 3D shape modeling tasks. Unless otherwise specified, the main configuration employed throughout the experiments—particularly for extrapolation to external datasets—was trained with 2,048 points per point cloud. However, experiments with 8,192 and 20,000 points were conducted and analyzed in section ~\ref{subsec:cloud_size} to ensure the ability to extract information and reconstruct large point clouds of automotive conditions.

\subsection{Training}
\label{subsec:training}

With the training datasets already prepared and organized, the next step was the training step. All models were trained on a single NVIDIA RTX 4090 GPU with 24 GB of VRAM, using a workstation equipped with a 13th Gen Intel\textsuperscript{\textregistered} Core\texttrademark{} i9-13900K and 64 GB of RAM. The model was optimized for 100 epochs using a batch size of 32 and an initial learning rate of $5\times10^{-4}$. Training employed the Adam optimizer~\cite{kingma2015adam}, and the reconstruction objective was defined as the Chamfer Distance (CD)~\cite{fan2017point} between the input and the reconstructed point sets. The architecture used in the majority of our experiments, and in particular for cross-dataset extrapolation to ModelNet10 and ModelNet40~\cite{wu20153d}, consisted of three encoder and three decoder blocks with $L= 1,024$. Unless otherwise specified, the model was trained with $M= 2,048$, which we found to provide a good trade-off between computational efficiency and reconstruction accuracy.

Prior to training, all point clouds were normalized to remove the effect of global translations and variations in scale. 
Given \({P} = \{ {p}_i \in \mathbb{R}^3 \}_{i=1}^N\), 
where \({p}_i\) represents the 3D coordinates of the \(i\)-th point, 
the centroid 
\(\bar{{p}} = \frac{1}{N}\sum_{i=1}^N {p}_i\)
(i.e., the mean position of all points) 
was first computed and subtracted from each point, effectively translating the cloud to the origin \((0,0,0)\). 
The cloud was then uniformly scaled so that its maximum distance from the origin equaled one:

\begin{equation}
p_i' = \frac{{p}_i - \bar{{p}}}{\max_{j}\|{p}_j - \bar{{p}}\|_2}, 
\quad 
i = 1,\dots,N,
\end{equation}
where \(\mathbf{p}_i'\) is the normalized point, $p_j$ denotes the \(j\)-th point in $P$, and \(\|\cdot\|_2\) denotes the Euclidean norm. This normalization ensured that all input point clouds shared a consistent spatial distribution, allowing the network to focus on structural features rather than global positioning.

The reconstruction objective was the Chamfer Distance ($CD$)~\cite{fan2017point} between the predicted reconstructed point cloud $R$ and the preprocessed input $X$, defined as:
\begin{equation}
CD (X, R) = 
\frac{1}{|{X}|}\sum_{{p}\in{X}} \min_{\hat{p}\in {R}} \|p-\hat{p}\|_2^2 
+ \frac{1}{|R|}\sum_{\hat{p}\in {R}} \min_{{p}\in X} \|\hat{{p}}-{p}\|_2^2 .
\end{equation}

While $CD$ is efficient and captures local proximity between point sets, it may overlook discrepancies in global structure. 
To address this, we also evaluated the reconstructions using the Earth Mover’s Distance ($EMD$) ~\cite{fan2017point}, defined as:
\begin{equation}
EMD ({X}, {R}) = 
\min_{\phi:{X}\to{{R}}} \frac{1}{|{X}|}\sum_{{p}\in{X}} \|{p}-\phi({p})\|_2 ,
\end{equation}
where $\phi$ denotes a bijection between point sets. 
Unlike $CD$, $EMD$ finds an optimal matching between the two sets, yielding a more faithful assessment of global shape similarity, at the expense of higher computational cost. 
This dual evaluation allows for a more comprehensive comparison with other models in the literature.

\subsection{Architecture Refinement Experiments}
\label{subsec:skip_relevance}

From the initial architecture and with the training dataset prepared, a series of studies were conducted to refine and optimize the model design. A systematic evaluation was performed to analyze the role of $S$ in the encoder–decoder structure, since these connections are typically present at every encoding layer. To this end, we consider various experiments to identify the optimal skip connection, such as the common skip connection setup $ss_1$ which connects all encoding layers with their corresponding decoding layer, $ss_2$ which connects only first encoding layer with last decoding layer, $ss_3$ which connects second encoder layer with second decoder layer or $ss_4$ which connects last encoding layer with first decoding layer. In order to ensure that the reconstruction relied solely on the information transmitted through $S$, the latent vector extracted by $E$ was replaced with a random vector before being replicated and passed to $D$.

\begin{figure}[t]
    \centering
    \begin{subfigure}[c]{0.30\linewidth}
        \centering
        \vspace*{0.5\linewidth}
        \includegraphics[width=\linewidth]{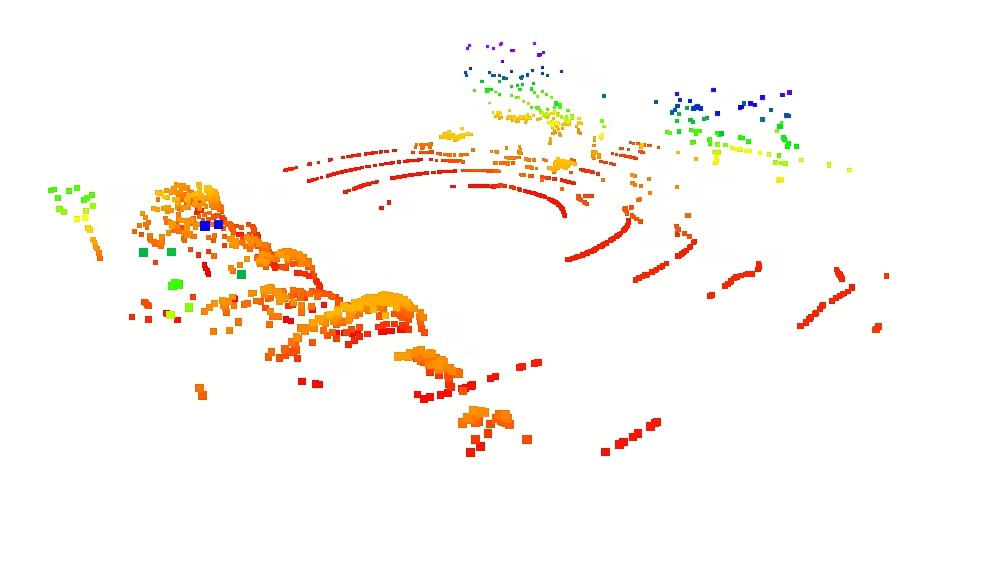}
        \caption{GT}
    \end{subfigure}%
    \hfill
    \begin{subfigure}[t]{0.30\linewidth}
        \centering
        \includegraphics[width=\linewidth]{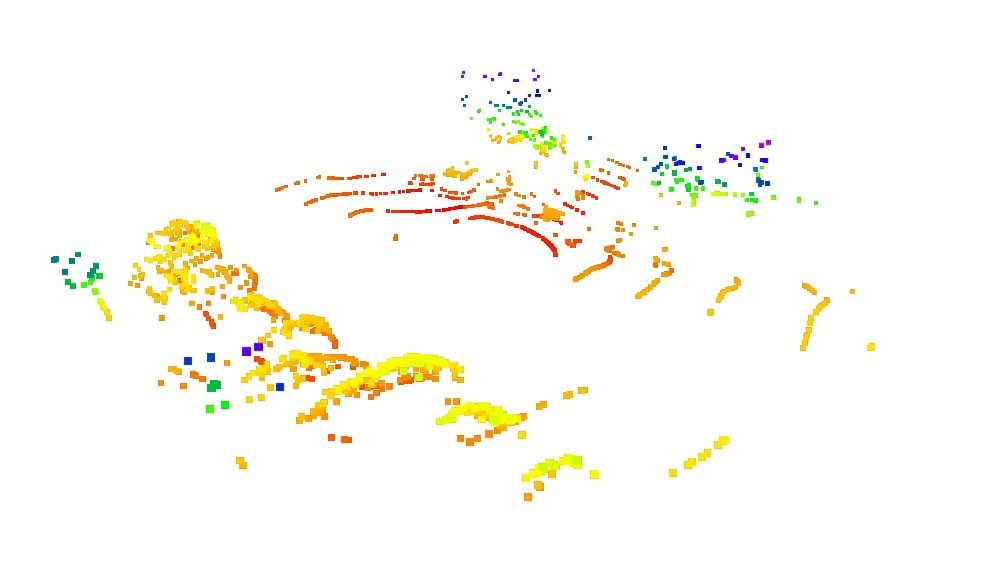}
        \caption{$ss_1$}
        \label{subf:all-skp}
        
        \includegraphics[width=\linewidth]{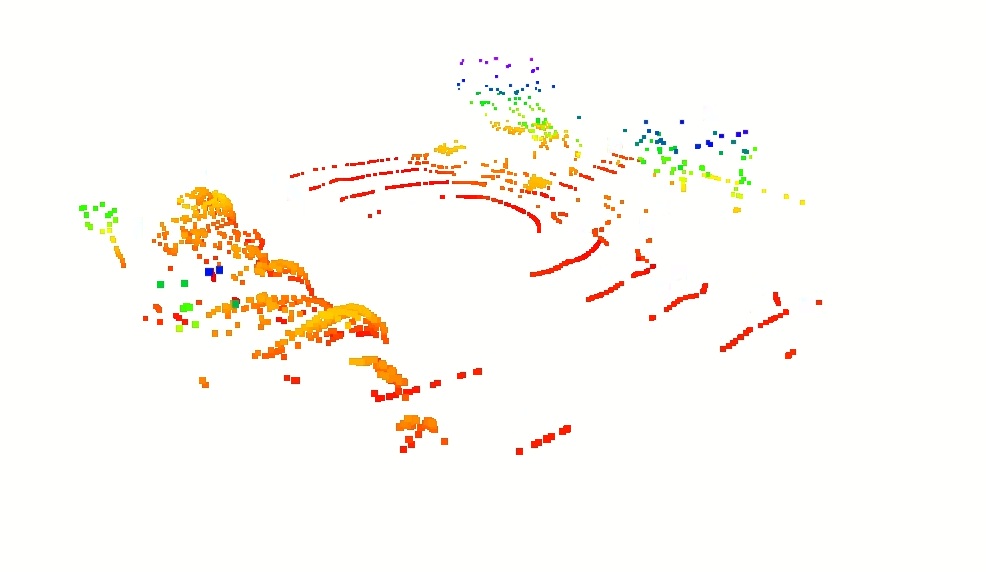}
        \caption{$ss_2$}
        \label{subf:outer-skp}
    \end{subfigure}%
    \hfill
    \begin{subfigure}[t]{0.30\linewidth}
        \centering
        \includegraphics[width=\linewidth]{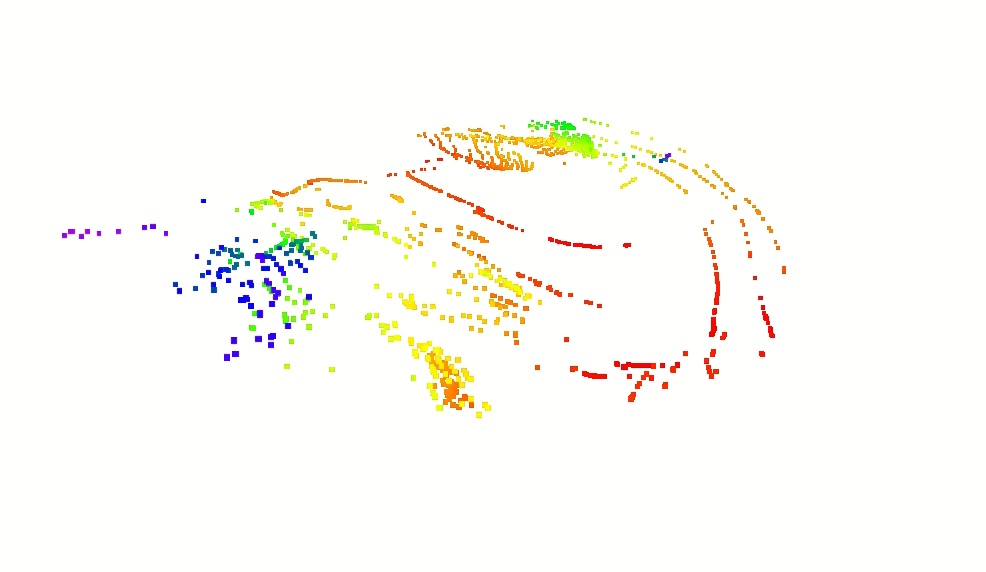}
        \caption{$ss_3$}
        \label{subf:middle-skp}
        
        \includegraphics[width=\linewidth]{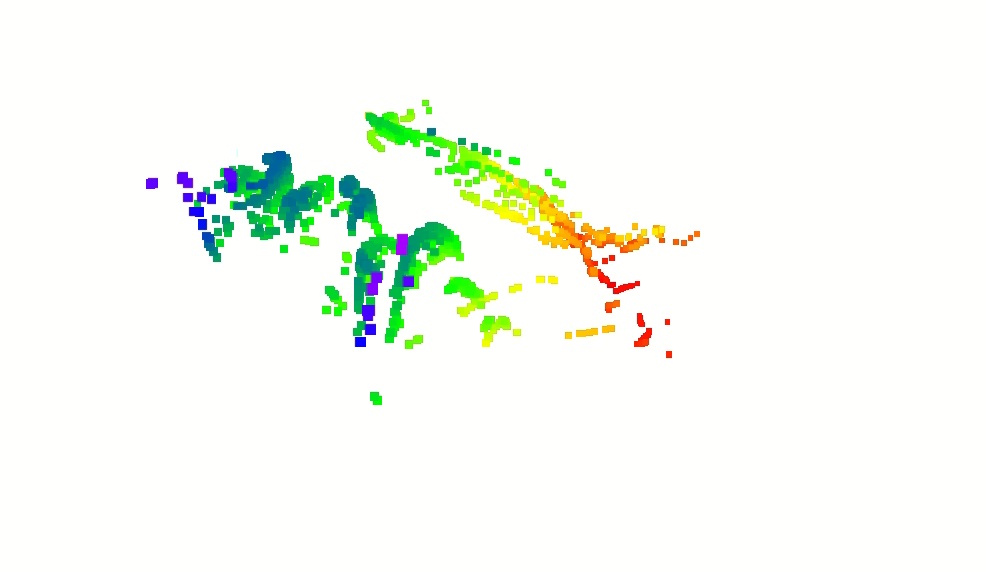}
        \caption{$ss_4$}
        \label{subf:inner-skp}
    \end{subfigure}
    \caption{Comparison of $R$ under different $S$ configurations Using a fixed random latent space. The color gradient along the height axis highlights how the reconstruction worsens with the depth of $S$ in the network.}
    \label{fig:1up_lsrandom}
    \vspace{-0.7em}
\end{figure}

The results, summarized in Fig. ~\ref{fig:1up_lsrandom} and Table ~\ref{tab:CD_EMD_skc}, show that $S$ in early layers of $E$ makes less relevant the encoded latent space. Reconstructions could still be achieved with reasonable quality even when the latent input to $D$ was random, indicating limited dependence on the latent space at early layers. However, in the case of $S$ in the last layer of $E$, reconstruction quality degraded significantly under the random latent space condition. However, when the random vector was replaced by the actual latent representation, reconstruction quality was restored, highlighting the critical importance of latent space information.

\vspace{-1pt}

\begin{table}[b]
    \centering
    \caption{Comparison of $CD$ and $EMD$ under Different $S$ Configurations.}
    \vspace{0.2cm}
    \resizebox{\columnwidth}{!}{
        \begin{tabular}{c c c c c}
        \textbf{Metric} & \textbf{$ss_1$} & \textbf{$ss_2$} & \textbf{$ss_3$} & 
        \textbf{$ss_4$}\\
        \hline 
        CD & 0.003164 & 0.000287 & 0.028953 & 0.210549\\
        EMD & 0.049724 & 0.014079 & 0.288271 & 0.498255\\
        \end{tabular}
    \label{tab:CD_EMD_skc}
    }
    \vspace{-1em}
\end{table}

These observations demonstrate two complementary findings. First, in point cloud autoencoders, acceptable reconstructions can be achieved with minimal encoding depth, relying primarily on outermost skip connections; however, in such cases the latent space carries little meaningful information and remains poor in quality. Second, when the objective is to obtain a rich and informative latent representation like in our case, the architecture can be designed with inner encoding layers while retaining only the final skip connection near the bottleneck. Table~\ref{tab:model_sizes} presents a comparative analysis across different methods and our skip connection configurations ($ss_1$, $ss_2$, $ss_3$, $ss_4$), using the experimental setup described in Section ~\ref{subsec:training}

\begin{table}
    \centering
    \caption{Comparative Results Across Different Methods and Our Implementations; Showing the impact of the number of skip connections on model parameters, overall size, and inference speed.}
    \vspace{0.2cm}
    \resizebox{\columnwidth}{!}{
    \begin{tabular}{r c c c}
        \textbf{Method} & \textbf{\# params (M)} & \textbf{Total Size (MB)} & \textbf{Inference Time (s)} \\
        \hline 
        AE-EM ~\cite{achlioptas2018learningrepresentationsgenerativemodels} & 39,650& 151.51 & 0.00270 \\

        Point-BERT~\cite{yu2022pointbertpretraining3dpoint} & 27,620 & 193.80 & 0.00450 \\
        FoldingNet~\cite{yang2018foldingnetpointcloudautoencoder} & 1,740 & 177.81 & 0.00360 \\
        Ours-$ss_1$ & 0,698 & 62.73 & 0.00162 \\
        Ours-$ss_2$ & 0,616 & 62.42 & 0.00159 \\
        Ours-$ss_3$ & 0,628 & 62.47 & 0.00156 \\
        Ours-$ss_4$ & 0,677 & 62.66 & 0.00161
    \end{tabular}
    \label{tab:model_sizes}
    }
    \vspace{-1em}
\end{table}

Finally, since the objective of this architecture is to obtain a higher-quality latent space, the remaining experiments were carried out using the version with only the innermost skip connection setup  $ss_4$.

\subsection{Evaluation with Varying Training Data}
\begin{figure}[b]
    \centering
    \includegraphics[width=\linewidth]{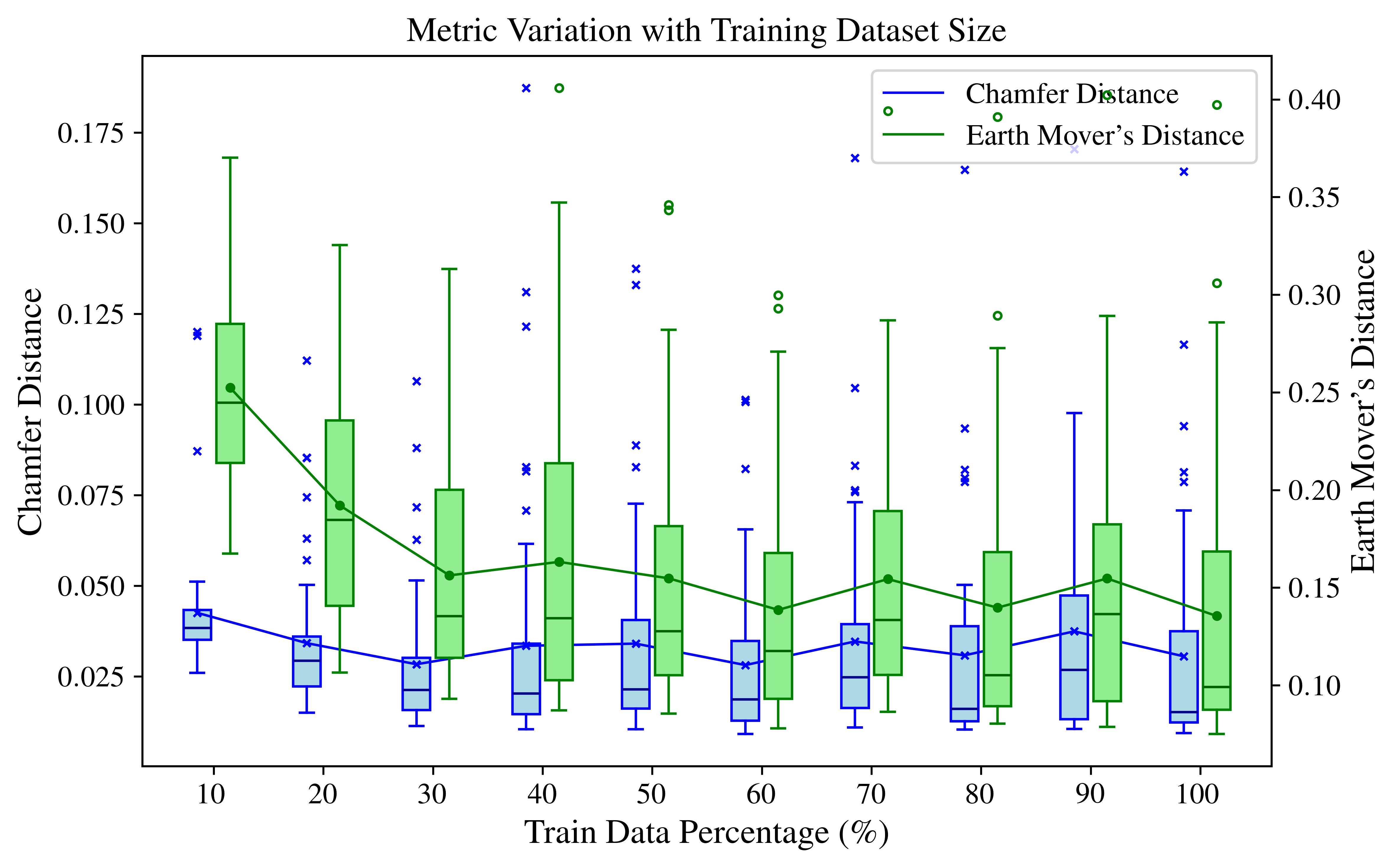}
    \caption{Evolution of Chamfer Distance ($CD$) and Earth Mover’s Distance ($EMD$) with varying training dataset size. Each point is based on 50 independent trainings per dataset size (500 in total). Line plots represent mean values (including outliers), while box plots depict the median and interquartile range.}
    \label{fig:ch_emd_varTrain}
    \vspace{-1em}
\end{figure}

To study the impact of data volume on our architecture, we generated a custom dataset composed of subsets of the original dataset (4460 point clouds). The dataset was divided into incremental proportions with a step size of 10\%. The objective of this procedure was to analyze the convergence behavior of the employed metric functions. To reduce the effect of stochastic variability, 50 independent training runs were performed for each custom dataset, resulting in a total of 500 training sessions. The outcomes of these experiments are summarized in Fig.~\ref{fig:ch_emd_varTrain}, where results are presented as a hybrid between a line plot and a box plot. Since the metrics do not share the same scale, each one is represented with its own axis. The box plots illustrate the distributions obtained, whereas the line plots indicate their mean values.

\subsection{Comparison Across Different Point Cloud Sizes}
\label{subsec:cloud_size}
Once accurate scene reconstruction was achieved using LiLa-Net, a study was conducted to assess the impact of $X$ size on reconstruction quality. The aim was to determine whether increasing the field of view and $M$ would introduce inaccuracies, particularly for regions farther from the origin, or create challenges in processing larger point clouds.

To this end, three models with identical architectures were trained using different sizes of $M$: 2,048 points, 8,192 points, and 20,000 points. Inference was then performed on the same sequences to compare reconstruction metrics, including $CD$ and $EMD$, as reported in Table~\ref{tab:points_models}.

\begin{table}[b]
    \centering
    \caption{Comparison of Reconstruction Performance and Processing Time for Different $M$ Values.}
    \vspace{0.2cm}
    \begin{tabular}{r c c c}
    \textbf{$M$} & \textbf{Inf. Time (ms)} & \textbf{$CD$ Loss} & \textbf{$EMD$ Loss}\\
    \hline 
    2,048 & 1.616 & $16.03 \times 10^{-5}$ & 0.126 \\
    8,192 & 5.612 & $9.50 \times 10^{-5}$ & 0.009 \\
   20,000 & 12.963 & $4.63 \times 10^{-5}$ & 0.008
    \end{tabular}
    \label{tab:points_models}
    \vspace{-1em}
\end{table}

\begin{figure}[b]
    \centering
    \begin{subfigure}[b]{0.32\linewidth}
        \centering
        \includegraphics[width=\linewidth]{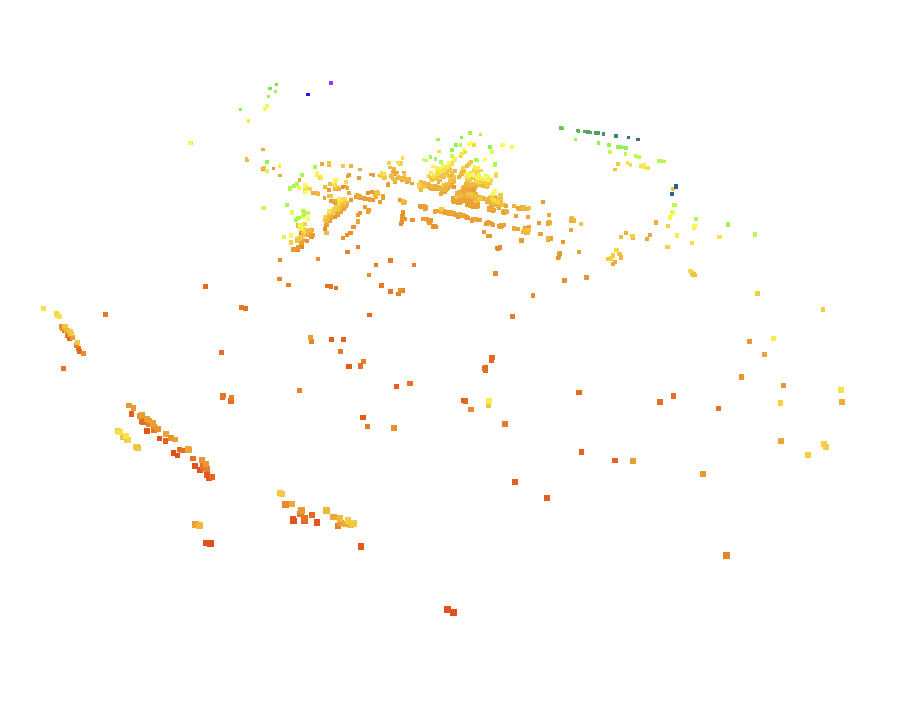}
        \caption{$M=$ 2,048}
    \end{subfigure}
    \begin{subfigure}[b]{0.32\linewidth}
        \centering
        \includegraphics[width=\linewidth]{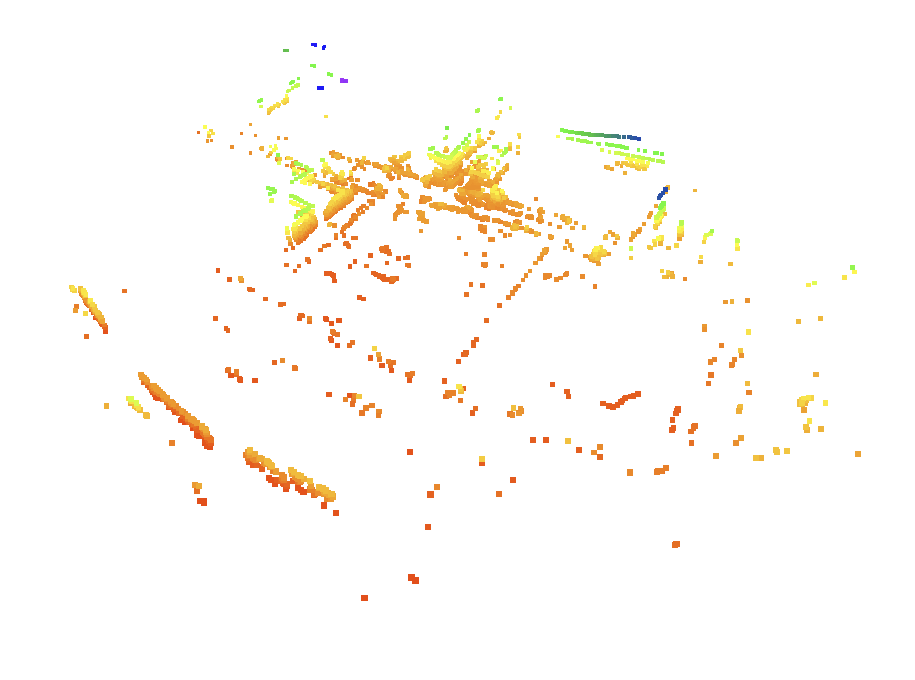}
        \caption{$M=$ 8,192}
    \end{subfigure}
    \begin{subfigure}[b]{0.32\linewidth}
        \centering
        \includegraphics[width=\linewidth]{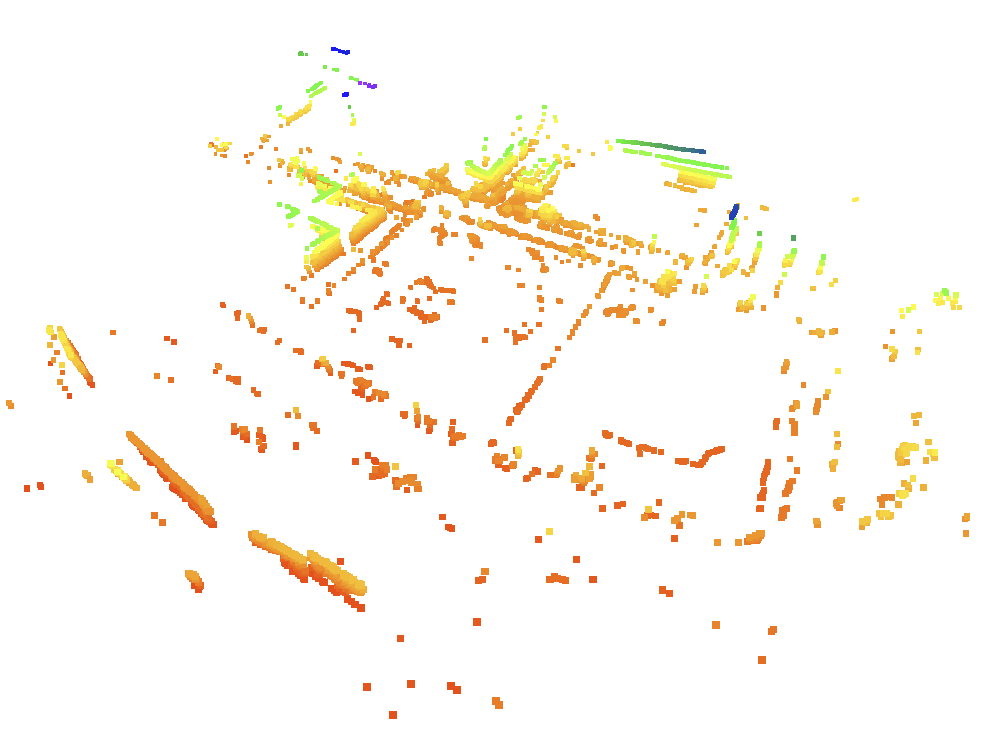}
        \caption{$M=$ 20,000}
    \end{subfigure}
    \caption{Reconstructed Point Clouds ($R_{3\times M}$) of the Same Scene Using Varying Input Cloud Sizes ($M$).}
    \label{fig:reco_points_models}
    \vspace{-1.1em}
\end{figure}

The results indicate that while input size has a significant effect on inference time, it does not compromise reconstruction accuracy. In fact, larger point clouds often achieve equal or slightly improved reconstruction quality due to the richer data representation. This improvement is also visually evident in Fig.~\ref{fig:reco_points_models}, which shows the reconstruction of the same scene with varying point cloud sizes.

\begin{table*}[h!]
    \centering
    \caption{Qualitative Comparison of Existing Point Cloud Models and Our Method on Our Platform Recording Dataset.}
    \vspace{0.2cm}
    \begin{tabular}{c c c c c}
        \textbf{GT} & \textbf{AE-EM~\cite{achlioptas2018learningrepresentationsgenerativemodels} } & \textbf{FoldingNet~\cite{yang2018foldingnetpointcloudautoencoder}} & \textbf{PointBert~\cite{yu2022pointbertpretraining3dpoint}} & 
        \textbf{LiLa-Net (Ours)} \\
        \hline \\
        \includegraphics[width=0.18\textwidth]{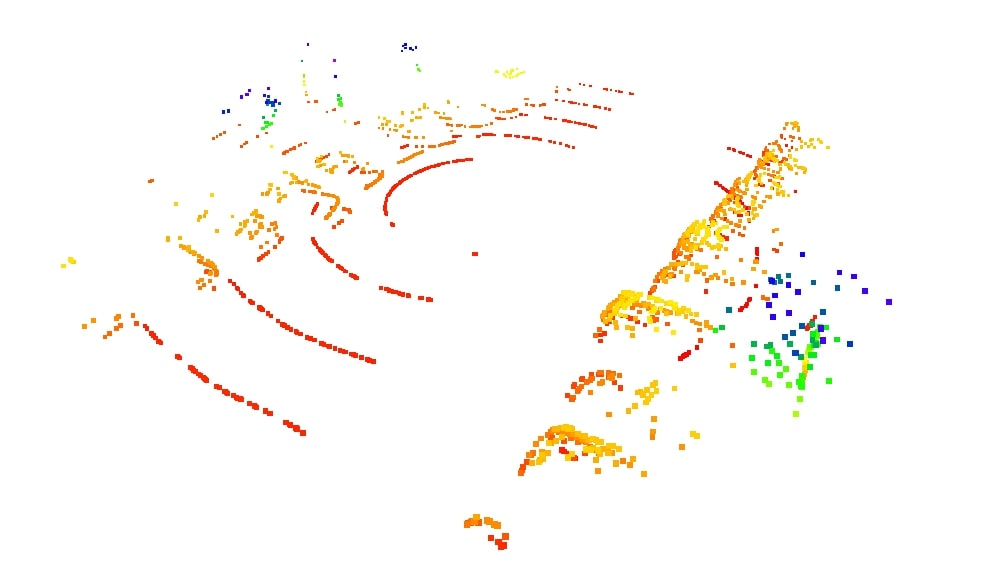} &
        \includegraphics[width=0.18\textwidth]{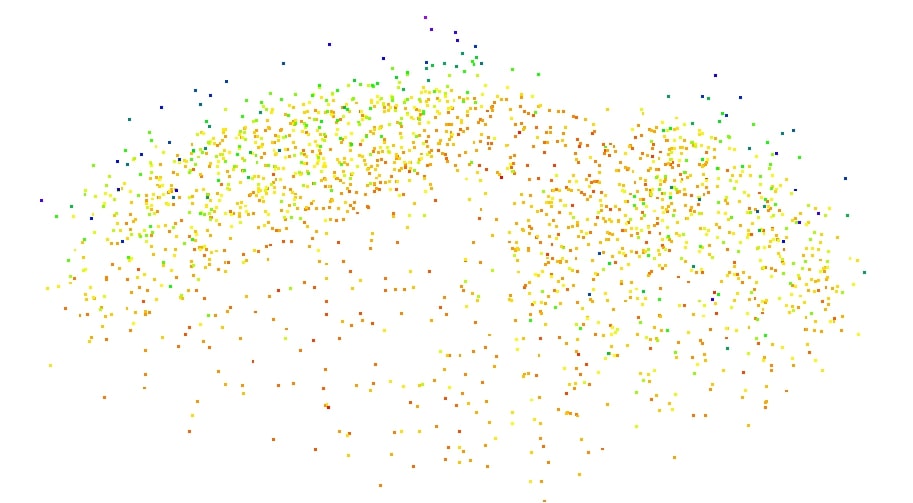} &
        \includegraphics[width=0.18\textwidth]{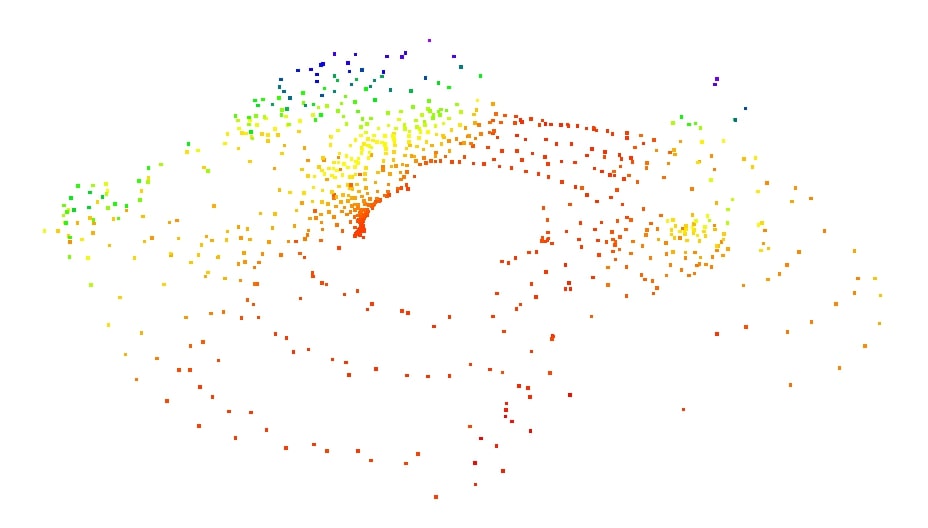} &
        \includegraphics[width=0.18\textwidth]{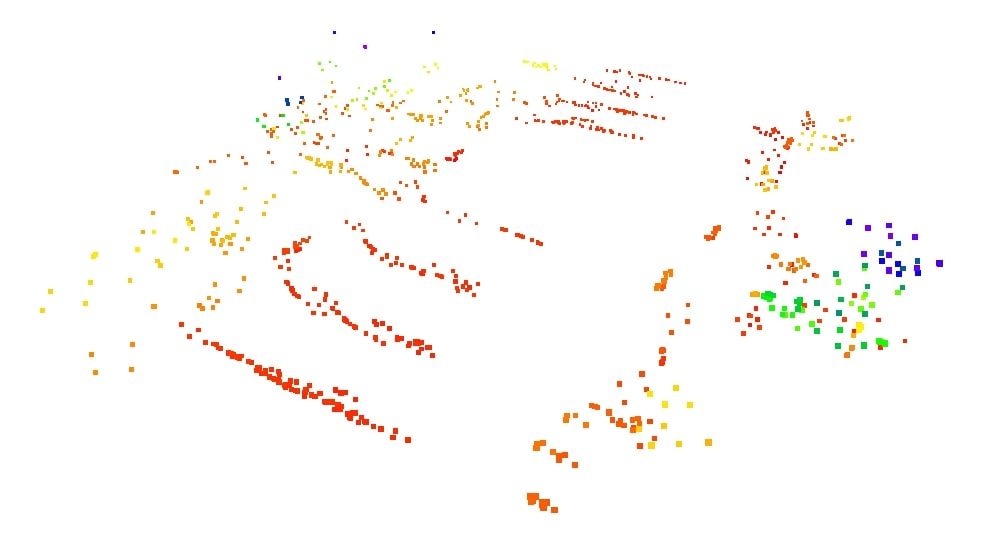} &
        \includegraphics[width=0.18\textwidth]{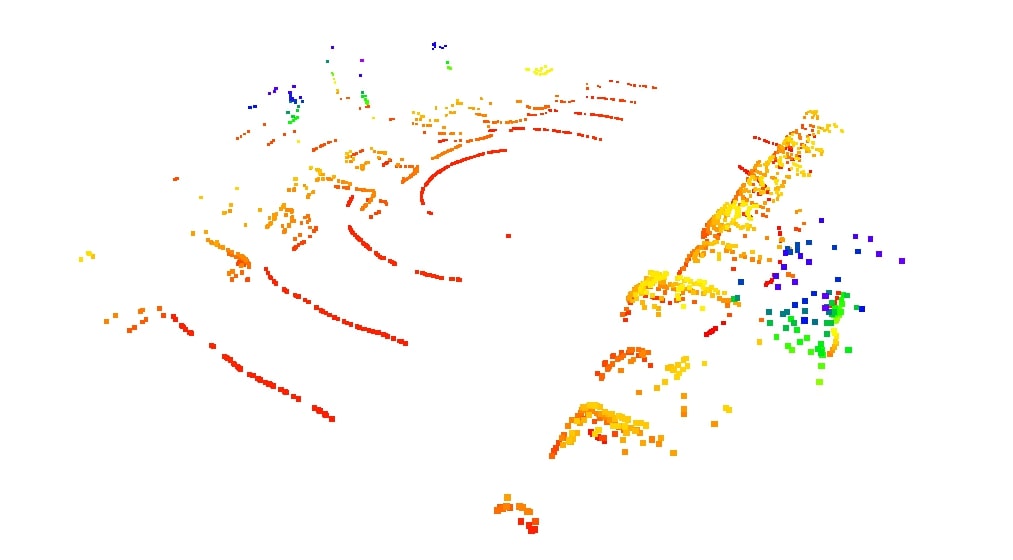}
        \\
        \\
        \includegraphics[width=0.18\textwidth]{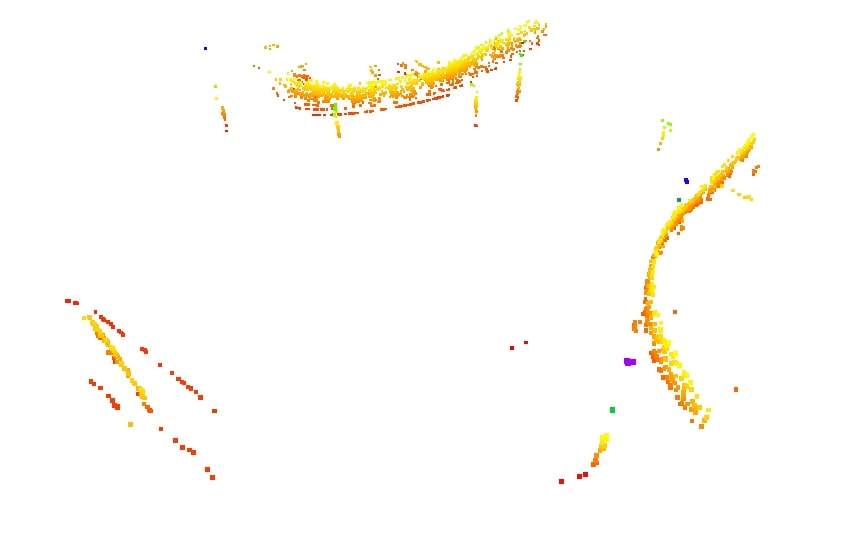} &
        \includegraphics[width=0.18\textwidth]{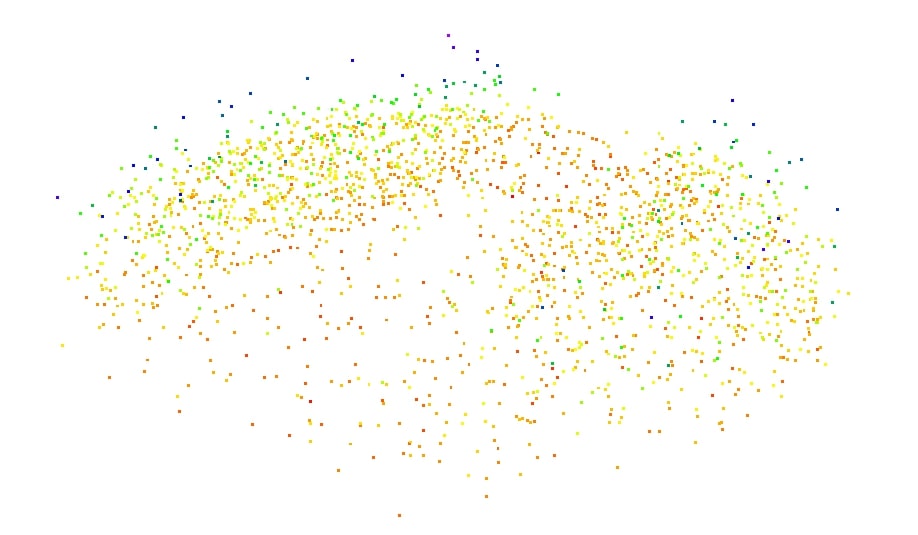} &
        \includegraphics[width=0.18\textwidth]{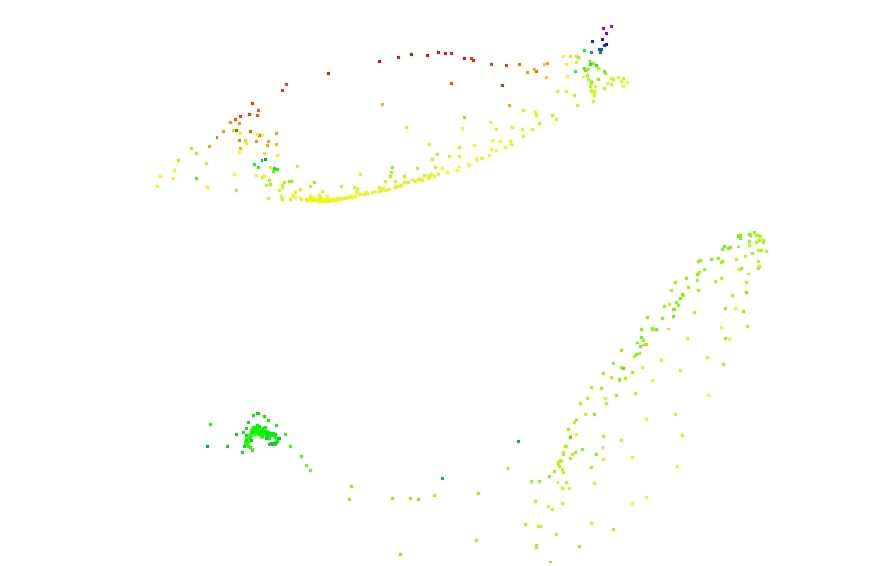} &
        \includegraphics[width=0.18\textwidth]{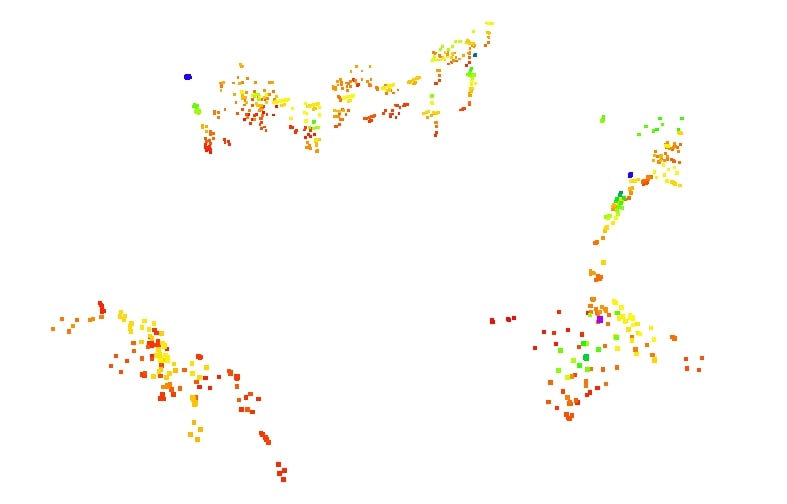} &
        \includegraphics[width=0.18\textwidth]{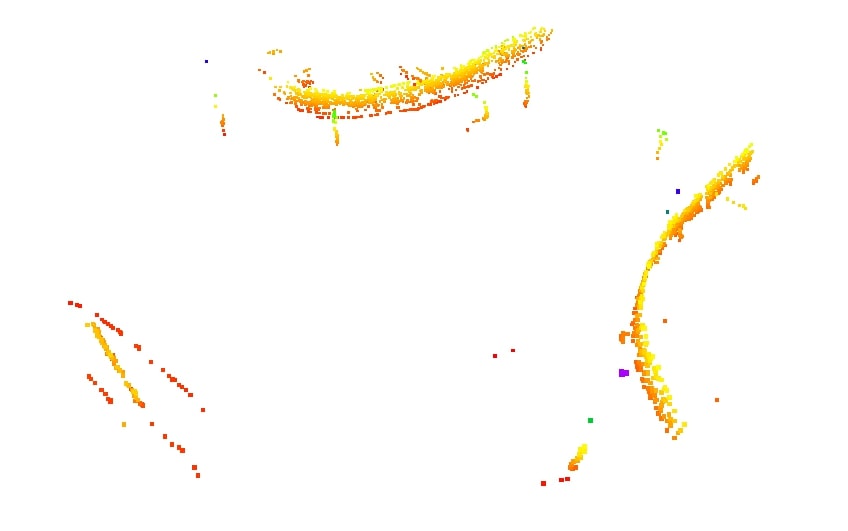} \\
        
    \end{tabular}
    \label{tab:hades_results}
\end{table*}

\subsection{Evaluation of Existing Point Cloud Models on Our Platform Recording}

To further validate the complexity of our platform recording, we tested several publicly available point cloud autoencoder repositories. Specifically, we evaluated the implementations from ~\cite{yang2018foldingnetpointcloudautoencoder, achlioptas2018learningrepresentationsgenerativemodels, yu2022pointbertpretraining3dpoint}. 

Our experiments revealed that the models from the Pytorch version of ~\cite{yang2018foldingnetpointcloudautoencoder, achlioptas2018learningrepresentationsgenerativemodels} failed to learn meaningful latent representations of our dataset, resulting in poor reconstructions and unstable training. In contrast, the implementation from~\cite{yu2022pointbertpretraining3dpoint} achieved relatively better training convergence and was able to reproduce point clouds more consistently. Nevertheless, the reconstructions remain far from satisfactory, underscoring the unique challenges posed by our platform recording compared to other standard datasets. 

A detailed comparison of reconstruction quality for each model is reported in Table~\ref{tab:hades_results}, where qualitative examples highlight the significant gap between existing approaches and the requirements for handling the complexity of our dataset.

Moreover, in Table~\ref{tab:CD_EMD_models}, we provide a quantitative evaluation based on $CD$ and $EMD$. The results confirm the qualitative observations reported in Table~\ref{tab:hades_results}. Specifically, both AE-EM~\cite{achlioptas2018learningrepresentationsgenerativemodels} and FoldingNet~\cite{yang2018foldingnetpointcloudautoencoder} exhibit high error values, reflecting their inability to capture the geometric complexity of our point clouds. PointBERT~\cite{yu2022pointbertpretraining3dpoint} achieves significantly lower distances, indicating more consistent reconstructions. However, our proposed model further reduces both CD and EMD by a large margin, highlighting its effectiveness in learning meaningful latent representations.

\begin{table}
    \centering
    \caption{Comparison of $CD$ and $EMD$ Across Point Cloud Models on Our Own Traffic Dataset.}
    \resizebox{\columnwidth}{!}{
        \begin{tabular}{c c c c c}
        \textbf{Metric} & \textbf{AE-EM~\cite{achlioptas2018learningrepresentationsgenerativemodels} } & \textbf{FoldingNet~\cite{yang2018foldingnetpointcloudautoencoder}} & \textbf{PointBert~\cite{yu2022pointbertpretraining3dpoint}} & 
        \textbf{LiLa-Net (Ours)}\\
        \hline 
        CD & 25.03 & 0.397 & 0.0109 & $16.03 \times 10^{-5}$ \\
        EMD & 5.430 & 7.493 & 0.136 & 0.012
        \end{tabular}
    \label{tab:CD_EMD_models}
    }
    \vspace{-1.1em}
\end{table}

\subsection{Cross-Dataset Extrapolation}

To evaluate the generalization ability of our autoencoder LiLa-Net beyond the domain of automotive LiDAR data, we investigated its performance on the widely used synthetic ShapeNet dataset~\cite{chang2015shapenet}. ShapeNet comprises over 50,000 3D CAD models spanning 55 object categories (e.g., airplanes, chairs, lamps), providing a diverse benchmark for 3D object representation learning.

We conducted extrapolation experiments by applying our model, trained exclusively on real-world LiDAR point clouds, directly to ShapeNet without any further fine-tuning. In these evaluations, we used the same processing pipeline as in the original domain, including normalization to unit sphere and fixed input size of 2,048 points, thereby assessing whether our learned latent representations can capture meaningful structure and facilitate accurate reconstruction in a drastically different setting.  

Qualitative results reveal that the model is capable of recovering fine-grained geometric details across a variety of object categories, even though it was never exposed to such shapes during training. As illustrated in Table~\ref{tab:modelnet40_reconstructions}, we present examples from six distinct classes of the ShapeNet dataset, showcasing the model’s ability to generalize across a wide spectrum of geometries.  

These findings highlight the ability of the proposed architecture not only to reconstruct data from familiar scenes, but also to generalize it to unknown and structurally different domains, reinforcing its potential for implementation in scenarios where data may be limited or domain-specific.

\subsection{Latent Space Evaluation}

To further assess the quality of the latent representations learned by our model, we conducted a classification experiment on the ModelNet10 and ModelNet40 datasets~\cite{wu20153d}, which are widely used benchmarks in 3D shape analysis. ModelNet10 consists of 10 object categories with 4,899 CAD models, while ModelNet40 contains 40 categories with 12,311 CAD models, providing a diverse collection of synthetic 3D objects suitable for evaluating the generalization ability of point cloud encoders.

Table~\ref{tab:modelnet40_acc} reports the classification accuracy obtained by training a linear SVM on the latent representations extracted by our model and several baselines from the literature.

\begin{table}
    \centering
    \caption{Comparison of Classification Accuracy on ModelNet10 and ModelNet40 Using Different Point Cloud Models.}
    \vspace{0.2cm}
    \resizebox{\columnwidth}{!}{
    \begin{tabular}{r c c c}
        \textbf{Method} & \textbf{SVM} & \textbf{Acc. ModelNet10 (\%)} & \textbf{Acc. ModelNet40 (\%)} \\
        \hline 
        PointNet~\cite{qi2017pointnet} & - & - & 86.20 \\
        3D-GAN~\cite{wu2017learningprobabilisticlatentspace} & \checkmark & 91.00 & 83.30 \\
        AE-EM~\cite{achlioptas2018learningrepresentationsgenerativemodels}& - & - & 85.70 \\
        VIP-GAN~\cite{han2018vipgan} & \checkmark & 94.05 & 91.98 \\
        FoldingNet~\cite{yang2018foldingnetpointcloudautoencoder} & \checkmark & 94.40 & 88.40 \\
        MAP-VAE~\cite{pang2022maskedautoencoderspointcloud} & \checkmark & 94.82 & 90.15 \\
        Point-Flow~\cite{yang2019pointflow}& \checkmark & 93.70 & 86.80 \\
        Point-BERT~\cite{yu2022pointbertpretraining3dpoint} & - & - & 93.20 \\
        MaskPoint~\cite{liu2022maskeddiscriminationselfsupervisedlearning}  & - & - & 93.80 \\
        Point- M2AE~\cite{zhang2022pointm2aemultiscalemaskedautoencoders} & \checkmark & - & 92.90 \\
        Point-MAE~\cite{pang2022maskedautoencoderspointcloud} & - & - & 93.80 \\
        MAE3D~\cite{jiang2023maskedautoencoders3dpoint}  & - & 95.50 & 90.60 \\
        Ours & \checkmark & 91.74 & 86.81
        
    \end{tabular}
    \label{tab:modelnet40_acc}
    }
    \vspace{-1em}
\end{table}

Some models specifically designed for classification or pre-trained on large-scale datasets such as ModelNet, ShapeNet, or ScanObjectNN achieve higher accuracy than ours. In contrast, our model was not pre-trained on any of these datasets and was originally developed for point cloud reconstruction rather than classification. Therefore, the purpose of this experiment is not to compete with SOA classification performance, but rather to provide a quantitative comparison of the expressiveness of the latent space.

\begin{table*}[htbp]
    \centering
    \caption{Qualitative Extrapolation Results on Shape-Net Dataset. Six object categories are shown, with ground truth (GT) and model predictions.}
    \vspace{0.2cm}
    \begin{tabular}{>{\centering\arraybackslash}m{1.1 cm} c c c c c c}
        & \textbf{Plane} & \textbf{Car} & \textbf{Sofa} & \textbf{Bed} & 
        \textbf{Bathtub} & \textbf{Printer} \\
        \hline \\
        \textbf{\vspace*{0.9cm}GT} & 
        \includegraphics[width=0.13\textwidth]{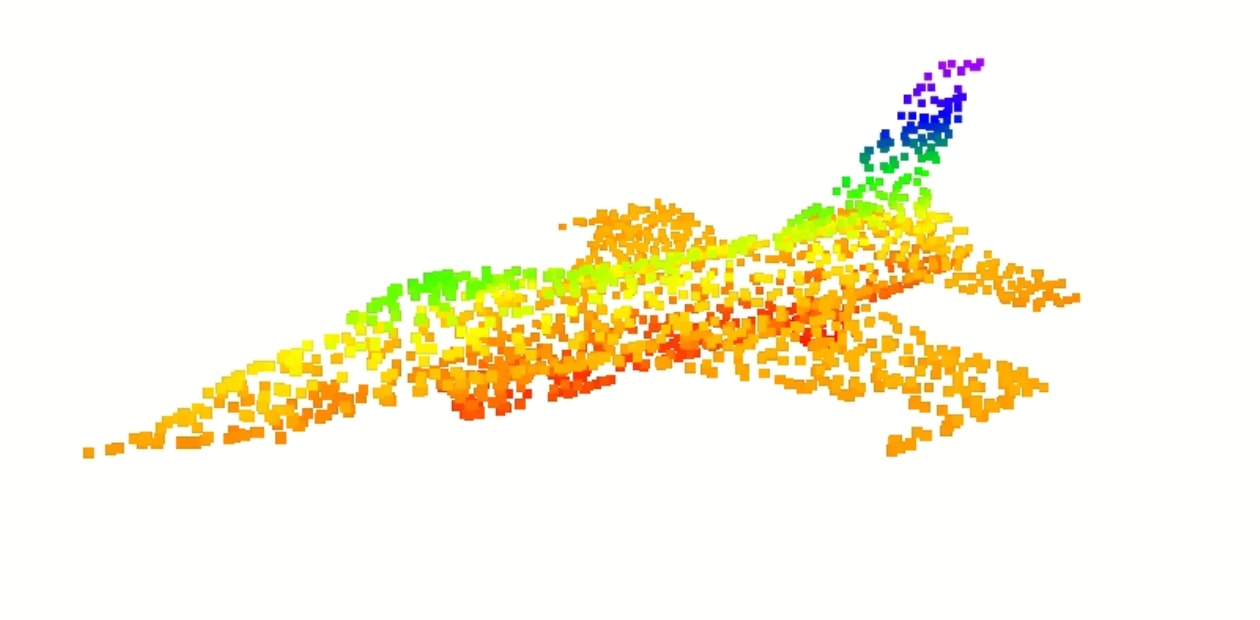} &
        \includegraphics[width=0.13\textwidth]{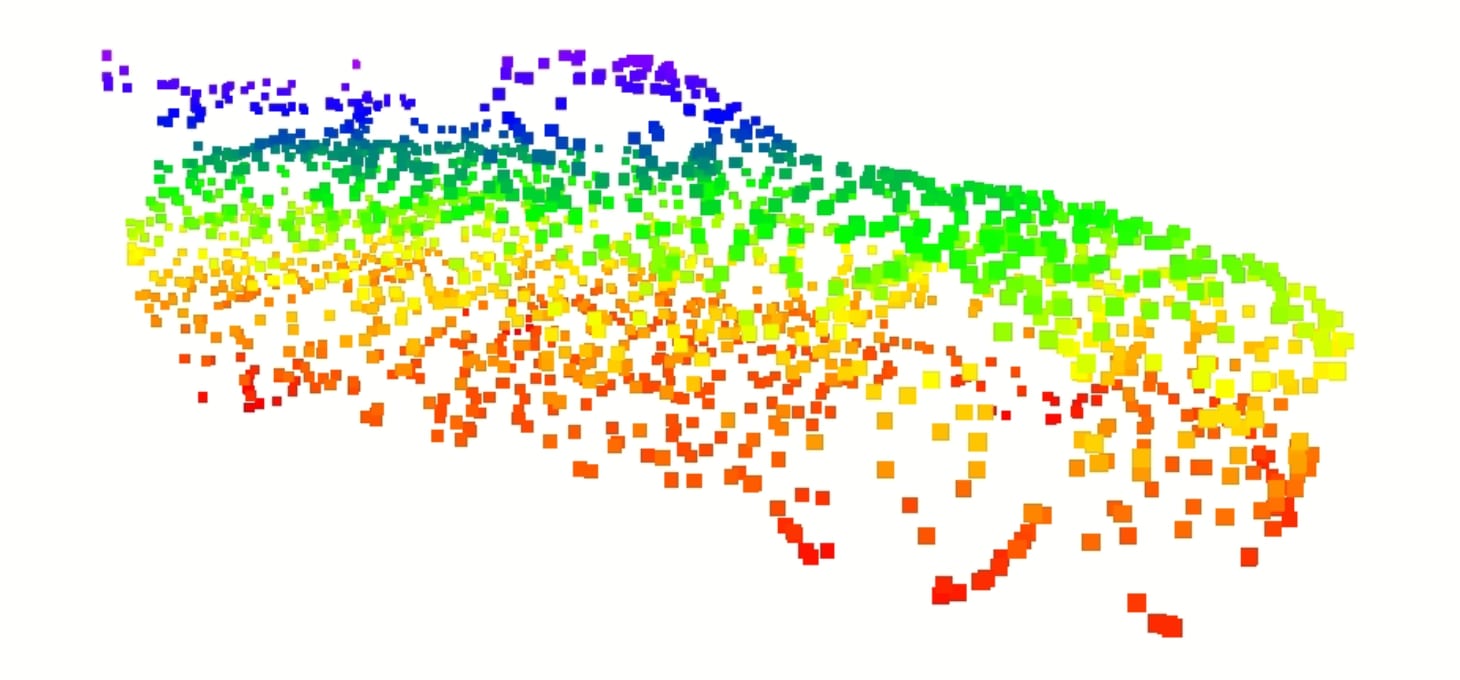} &
        \includegraphics[width=0.13\textwidth]{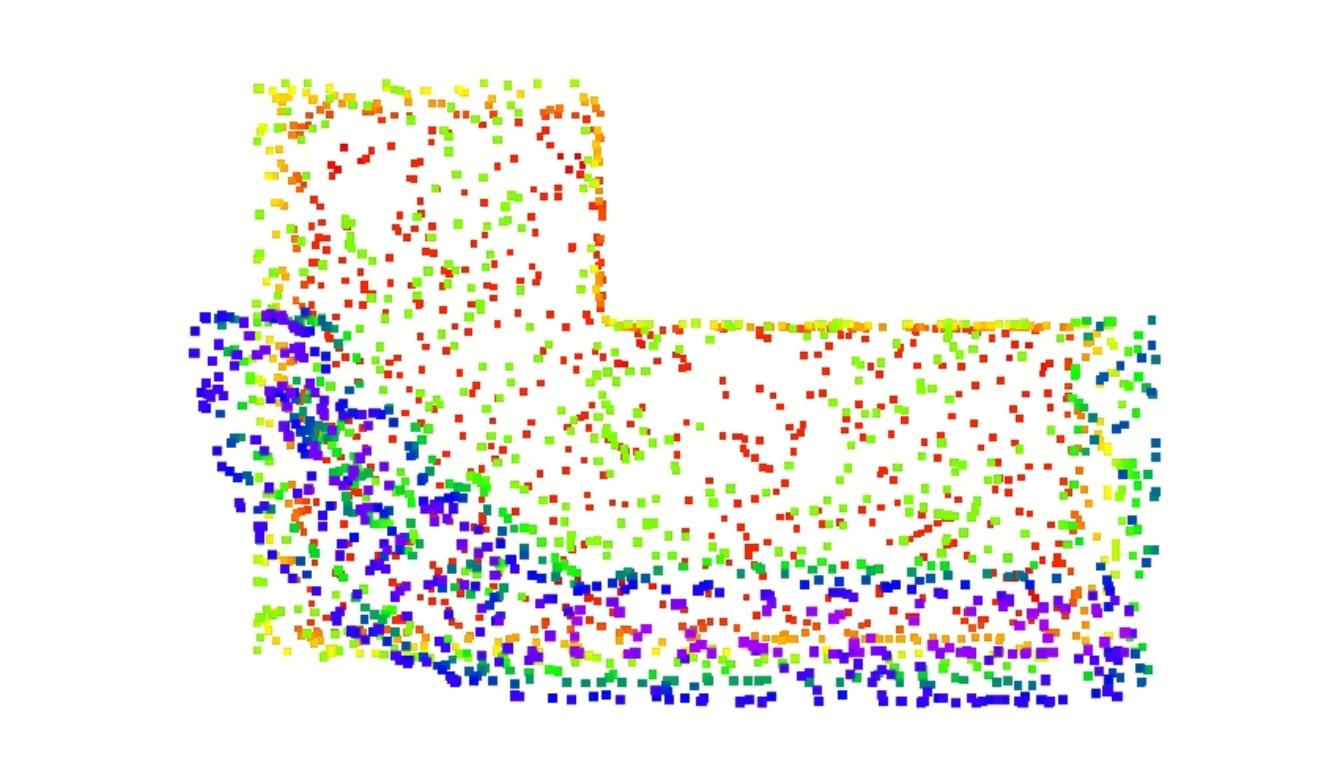} &
        \includegraphics[width=0.13\textwidth]{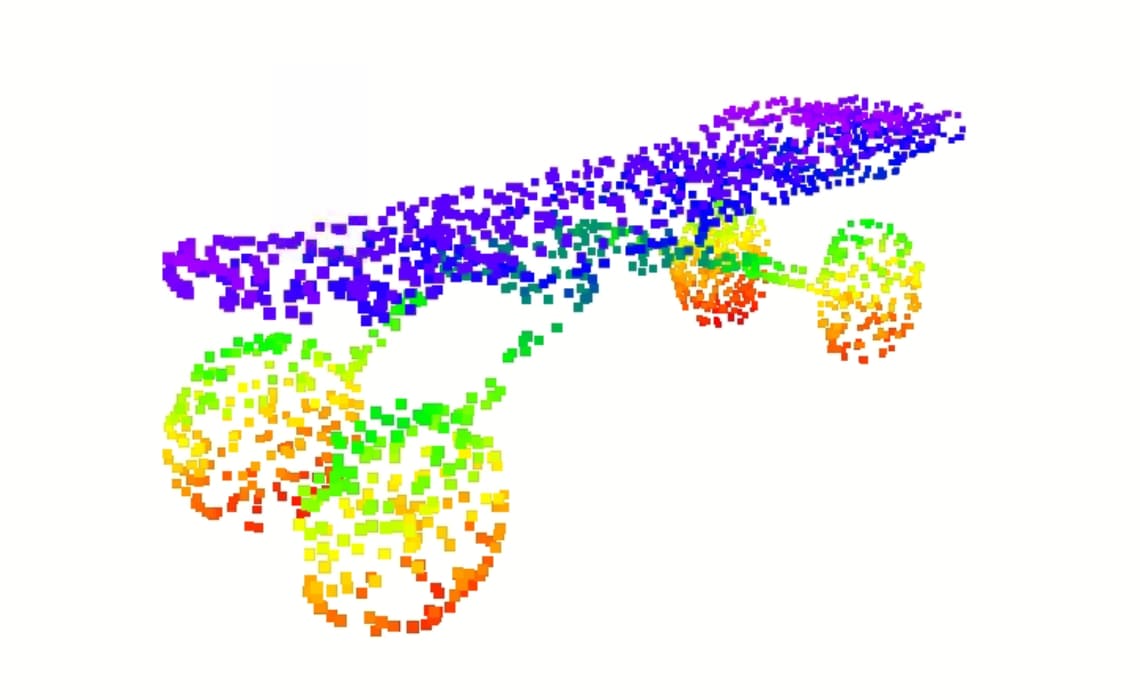} &
        \includegraphics[width=0.13\textwidth]{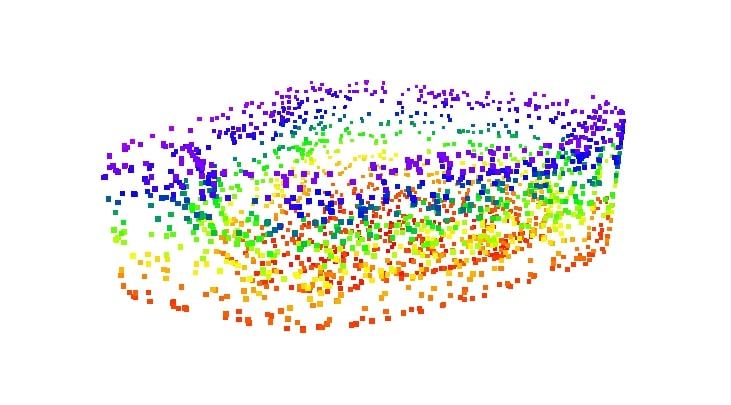} &
        \includegraphics[width=0.13\textwidth]{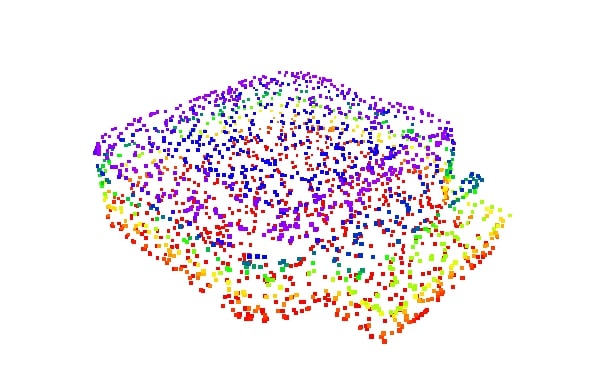} 

        \\
        \\
        
        \textbf{\vspace*{0.9cm}{LiLa$-$Net}} &
        \includegraphics[width=0.13\textwidth]{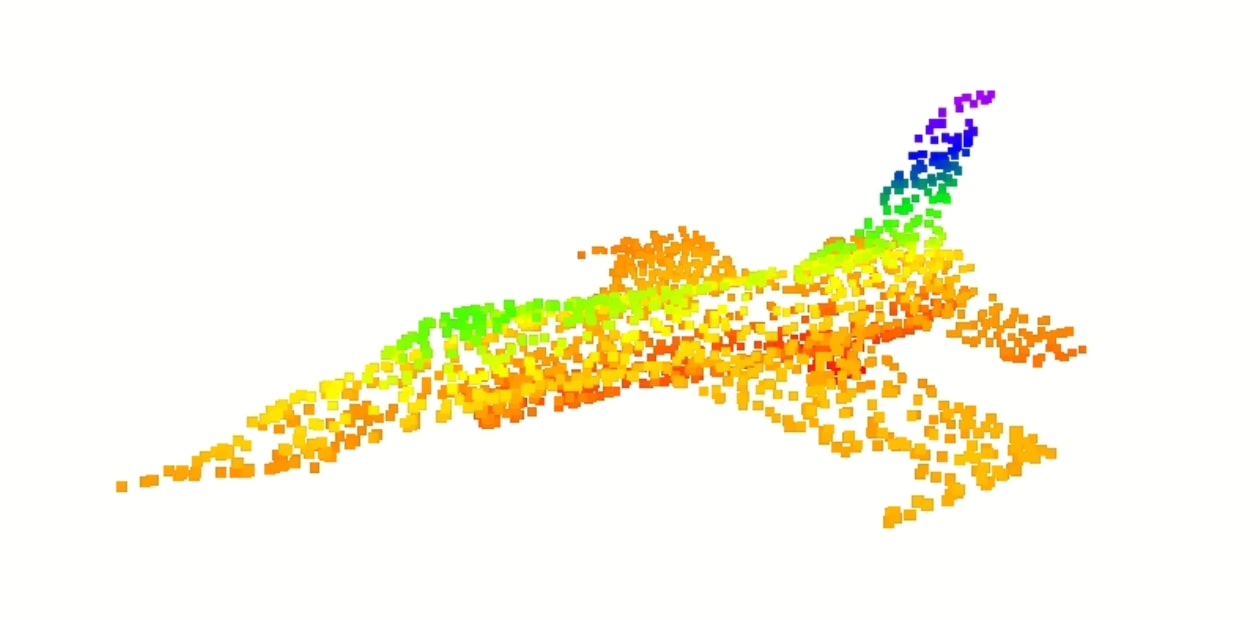} &
        \includegraphics[width=0.13\textwidth]{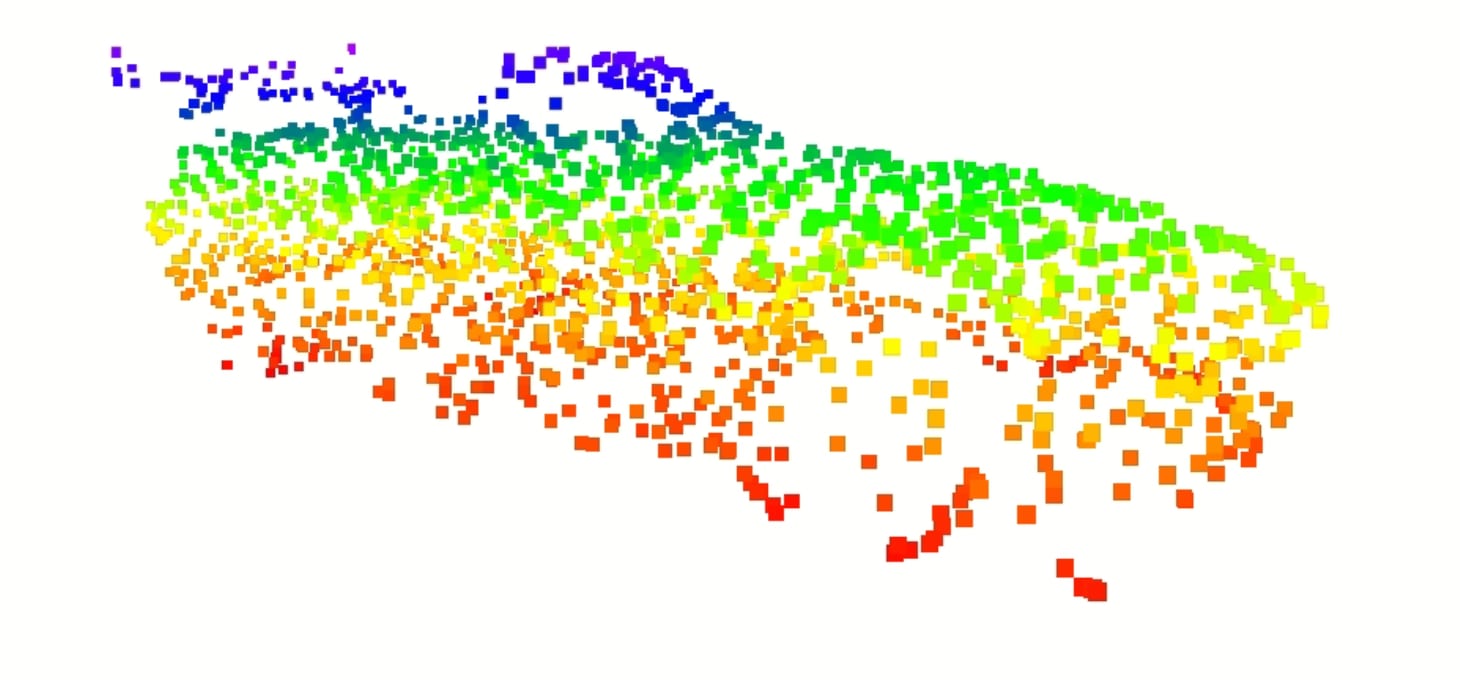} &
        \includegraphics[width=0.13\textwidth]{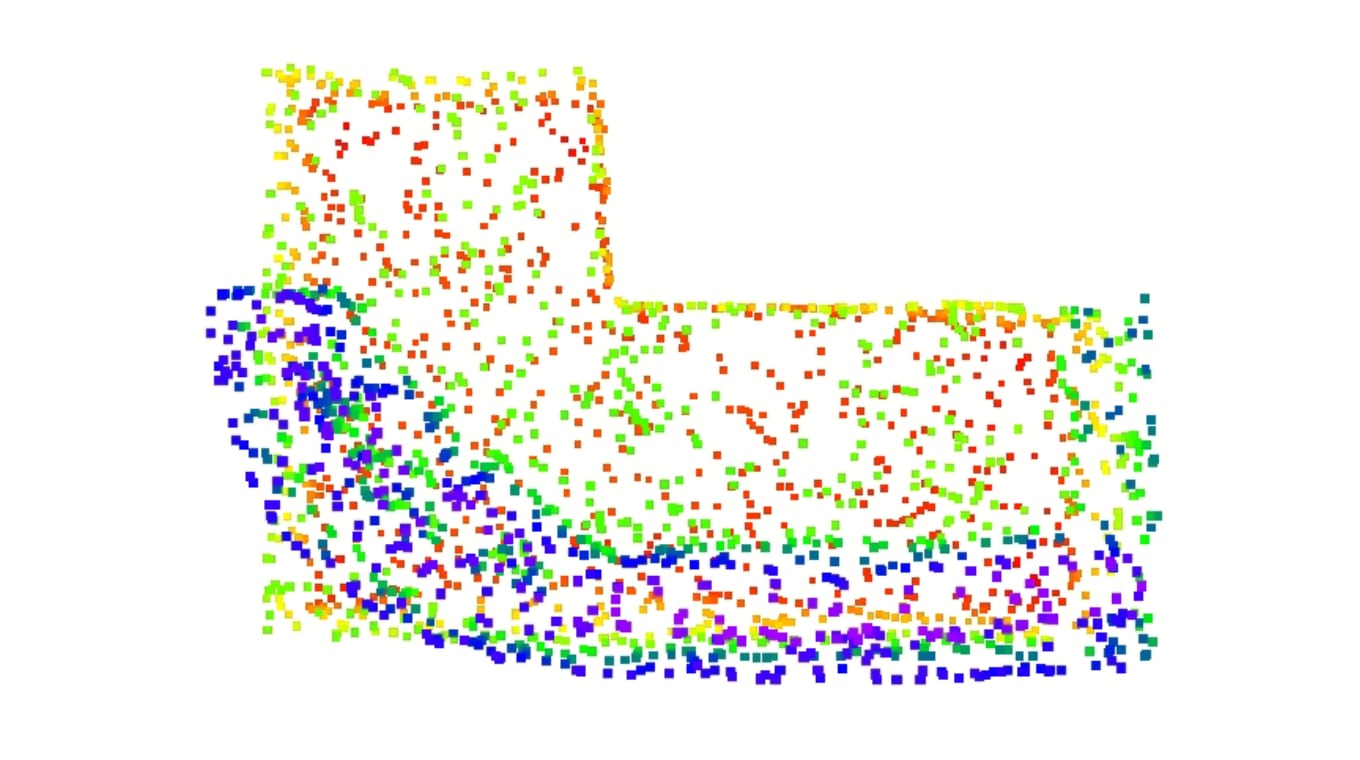} &
        \includegraphics[width=0.13\textwidth]{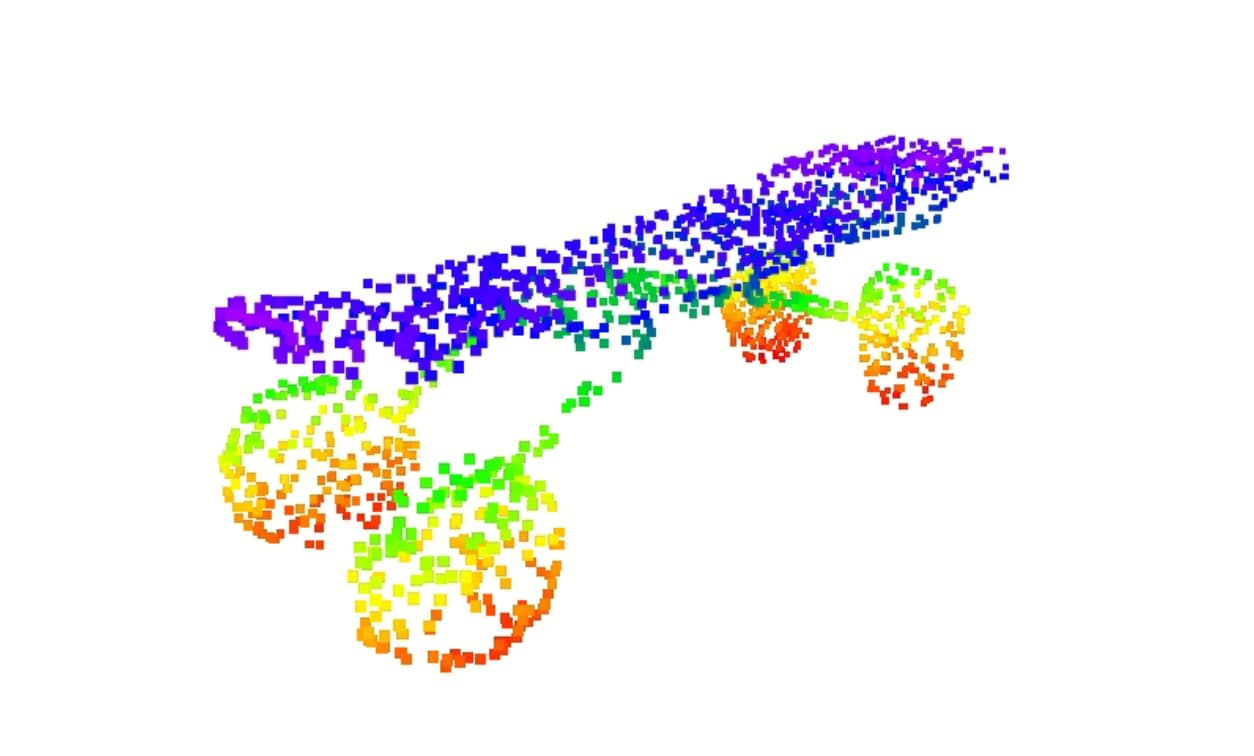} &
        \includegraphics[width=0.13\textwidth]{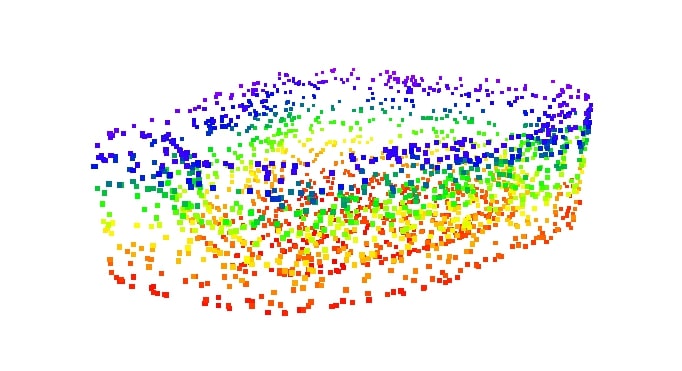} &
        \includegraphics[width=0.13\textwidth]{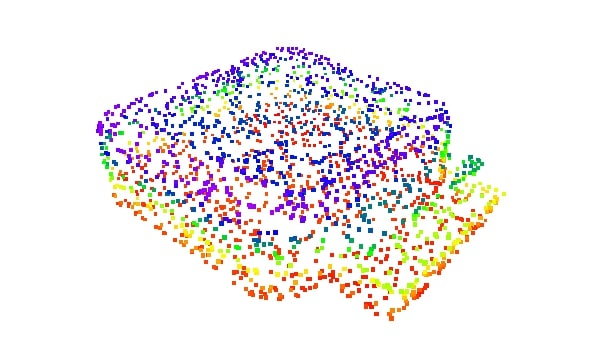} \\
        
    \end{tabular}
    \label{tab:modelnet40_reconstructions}
\end{table*}

\section{Conclusions}\label{sec:conclusions}
This work presented the implementation of a robust domain-shift autoencoder for point cloud generation with skip connections across different layers of the architecture to obtain a representative latent space from which the original data can be reconstructed. The proposed design achieved strong results in point cloud reconstruction, as demonstrated both by qualitative evaluations and by quantitative metrics such as accuracy, Earth Mover's Distance ($EMD$) and Chamfer Distance ($CD$). Experiments conducted on diverse datasets and point cloud representations further confirmed the reliability and precision of the approach.
In summary, the results confirm that the proposed architecture provides a robust and effective framework for 3D point cloud reconstruction, consistently producing accurate and high-quality representations across diverse datasets.

\section*{Acknowledgment}
The work carried out in this paper is part of the EcoMobility project. This project is supported by the CHIPS Joint Undertaking and its members, including top-up funding from the national authorities of Türkiye, Spain, the Netherlands, Latvia, Italy, Greece, Germany, Belgium, and Austria under grant agreement number 101112306. Co-funded by the European Union.

\bibliographystyle{ieeebib}
\bibliography{refs.bib}

\end{document}